\documentclass[runningheads]{llncs}
\usepackage[T1]{fontenc}
\usepackage{graphicx} 
\usepackage{amsmath}
\usepackage{amsfonts}
\usepackage{multirow}
\usepackage{booktabs}
\usepackage{xcolor}
\usepackage{multirow}

\begin{document}
\title{A Practice of Post-Training on Llama-3 70B with Optimal Selection of Additional Language Mixture Ratio}
\titlerunning{Post-Training on Llama-3 with Optimal Additional Language Mixture Ratio}
%
\author{Ningyuan Xi\inst{1,2}\thanks{\;This work is done during Ningyuan's internship in Geely.}\orcidID{0009-0008-7529-4660} \and
Yetao Wu\inst{1}\orcidID{0000-0003-3668-6414} \and Kun Fan\inst{1}\orcidID{0009-0001-8394-9507} \and Teng Chen\inst{1}\orcidID{0009-0000-3604-5313} \and Qingqing Gu\inst{1}\orcidID{0009-0007-6872-8910} \and Luo Ji\inst{1}\thanks{\;Corresponding author.}\orcidID{0000-0002-2484-5345}
}
\authorrunning{N. Xi et al.}
%
\institute{Geely AI Lab
\email{\{Yetao.Wu,Kun.Fan1,Teng.Chen2,Qingqing.Gu3,Luo.Ji1\}@geely.com}
\and Beihang University \\
\email{21373102@buaa.edu.cn} 
}
\maketitle              
\begin{abstract}
Large Language Models (LLM) often need to be Continual Pre-Trained (CPT) to obtain unfamiliar language skills or adapt to new domains. The huge training cost of CPT often asks for cautious choice of key hyper-parameters such as the mixture ratio of extra language or domain corpus. However, there is no systematic study that bridges the gap between the optimal mixture ratio and the actual model performance, and the gap between experimental scaling law and the actual deployment in the full model size. In this paper, we perform CPT on Llama-3 8B and 70B to enhance its Chinese ability. We study the optimal correlation between the Additional Language Mixture Ratio (ALMR) and the Learning Rate (LR) on the 8B size which directly indicates the optimal experimental setup. By thorough choice of hyper-parameter, and subsequent fine-tuning, the model capability is improved not only on the Chinese-related benchmark but also in some specific domains including math, coding, and emotional intelligence. We deploy the final 70B version of LLM on a real-life chat system which obtains satisfying performance.
\end{abstract}

\section{Introduction}


As part of post-training, Continual Pre-Training (CPT) is a widely used technique to enhance the fundamental capability of a pre-trained Large Language Model (LLM) in case the pre-trained corpus is insufficient. Purposes of CPT can generally be classified into two categories: to expand LLM's proficiency on some extra language~\cite{ke2023continuallearningnaturallanguage,cui2023ChineseLLAMA,chen2024effectiveefficientcontinualpretraining}, and to enhance domain-specific knowledge such as code, math and law~\cite{gururangan2020dontstoppretrainingadapt,ma2023ecomgptctcontinualpretrainingecommerce,wu2024llamaproprogressivellama,liang2024taskorientedindomaindata}. However, similar with the pre-training stage, CPT also suffers from enormous GPU consumption and tedious time cost. As a result, the optimally experimental configuration of CPT is generally unclear since a full grid search is intractable. Furthermore, CPT often results in catastrophic forgetting phenomena with inappropriate hyper-parameters~\cite{ke2023continuallearningnaturallanguage}. These issues call for thorough studies of CPT on the training corpus preparation and mixing, the hyper-parameter determination, as well as the selection and evaluation of typical downstream tasks.

Scaling law studies could provide insightful guidance on the CPT practice, which shares some basic ideas with scaling laws of pre-training~\cite{Kaplan2020OpenAIscalinglaw,Hoffmann2022DeepmindScalingLaw}. Among these efforts, D-CPT Law~\cite{que2024dcptlawdomainspecificcontinual} proposes a scaling law of the optimal mixture ratio between the general corpus and the domain corpus, as a function of different dataset sizes and model sizes. A subsequent study~\cite{gu2024cmrscalinglawpredicting} further defines the maximum mixture ratio within the feasible region as Critical Mixture Ratio (CMR), and discovers its correlation with loss and training tokens. These studies provide some feasible manners to set up the CPT experiments, but also subject to some shortcomings. First, almost all of such studies indicate the model performance with the pre-training loss, which, however, might not be equalized with the task performance, as indicated by our CPT experiments. Second, CMR assumes one can choose the largest feasible mixture ratio, which might be suboptimal when evaluating downstream task performance. Finally, their experiments are often conducted on relatively small model sizes (up to 4B), sometimes resulting in a gap with some industry-level LLM applications with a size of 7B or even larger.




On the other hand, there is also a lack of systematic studies on subsequent fine-tuning and downstream task performance, as a function of the CPT configuration. While it is observed that there sometimes exists the `alignment tax' on some specific domains comparing the fine-tuned LLM with its pre-trained version~\cite{cai-etal-2023-improving} (Table~\ref{tab:degradation}), such performance degradation can be alleviated by intentionally designed CPT on the domain-enhanced corpus, and re-gain or further improve the skills by the instruction fine-tuning after CPT~\cite{que2024dcptlawdomainspecificcontinual}. The overall performance of the entire pipeline requires the comprehensive consideration of CPT and fine-tuning experimental settings, which is seldom discussed.

\begin{table}[h!]
\caption{Examples of performance degradation of fine-tuned LLM comparing with pre-trained version. Results of HellaSwag, Winogrande, GPQA and MUSR are obtained from the open-LLM-leaderboard (\url{https://huggingface.co/spaces/open-llm-leaderboard/}). Results of LCSTS with the corresponding Chinese token faction (\%Chinese) are evaluated by ourselves.}
    \label{tab:degradation}
    \begin{center}
    \setlength\tabcolsep{3pt}
    \begin{tabular}{lcccccc}
        \toprule
        Model & HellaSwag & Winogrande & GPQA & MUSR & LCSTS & \%Chinese \\ 
        \midrule
        Llama-3-8B & 82.09 & 77.35 & 7.38 & 6.24 & 9.64 & 27.84 \\
        Llama-3-8B-Instruct & \textcolor{red}{78.55} & \textcolor{red}{74.51} & \textcolor{red}{5.70} & \textcolor{red}{5.40} & \textcolor{red}{0.34} & \textcolor{red}{0.09} \\
        \midrule
        Llama-3-70B & - & - & 19.69 & 16.01 & 7.19 & 34.49 \\
        Llama-3-70B-Instruct & - & - & \textcolor{red}{4.92} & \textcolor{red}{10.92} & \textcolor{red}{0.37} & \textcolor{red}{2.10} \\ 
        \bottomrule
    \end{tabular}
\end{center}
    
\end{table}

In this paper, we focus on the misaligned problem of Llama-3-instruct~\cite{grattafiori2024llama3herdmodels} on a specific language (Chinese) capability, and design an efficient and effective experiment to enhance the Llama-3 performances on not only Chinese benchmarks but also overall skills. The original pre-trained Llama-3 8B and 70B have moderately Chinese performance but often fail to respond to follow Chinese instructions after fine-tuning. To address this issue, we first study the optimal experimental configuration of early CPT practices on 8B, then conduct the formal 8B and 70B experiments on the CPT, Supervised Fine-tuning (SFT)~\cite{Ouyang2022ChatGPT} and Direct Preference Optimization (DPO)~\cite{Rafailov2023DPO} pipeline.  We first propose the variable of the Additional Language Mixture Ratio (ALMR) within the CPT corpus, then study the efficient frontier between ALMR and learning rate (LR), with the proxy of validation loss, benchmark performance and the training dataset size fixed. The optimal pairwise choice of ALMR and learning rate are then applied to the formal 8B and 70B CPT studies, as a basis of subsequent SFT and DPO training. We find that model performances are improved for both model sizes, on Chinese-related benchmarks, as well as other domain-specific benchmarks such as math, coding, and emotional intelligence. The finally enhanced LLM is deployed on a working companion with end-to-end chat quality substantially improved. We argue that appropriate choices of ALMR and LR can substantially help the LLM effectively adapt to unfamiliar languages or domains and might even inspire more generalized capabilities. Ablation studies also indicate our approach performs better than solely CPT or SFT using the same training datasets. To summarize, the main contributions of this paper include the following:



\begin{itemize}
\item We propose an optimal correlation between ALMR and LR with continual pre-training on Llama-3 8B and 1.7T training tokens, both of which are the largest size among such types of studies. 
\item We perform CPT, SFT, and DPO on Llama-3 8B and 70B, and obtain strengthened downstream performances by optimal CPT setup.



\item We deploy the model on a chatbot application on empathetic conversations, with human-annotated performance and emotional benchmarks improved.
\end{itemize}


\section{Related Work}
\label{sec:recent_work}

\subsection{Continual Pre-Training}





Continually Pre-Train (CPT) is generally employed to enhance LLM into two types of tasks, to expand the linguistic ability of LLM to a new language~\cite{ke2023continuallearningnaturallanguage,cui2023ChineseLLAMA,chen2024effectiveefficientcontinualpretraining}, or adapt to specific domains or tasks~\cite{gururangan2020dontstoppretrainingadapt,ma2023ecomgptctcontinualpretrainingecommerce,wu2024llamaproprogressivellama,liang2024taskorientedindomaindata}. For example, Cui et al.,~\cite{cui2023ChineseLLAMA} developed a method to enhance LlaMA-2 13B's Chinese language capabilities by extending its vocabulary with 20,000 Chinese tokens followed by CPT on Chinese corpora with LoRA. This approach demonstrates how CPT can be applied to expand a model's linguistic abilities. The mixture ratio of the Chinese corpus to the general corpus is 2:8.  Linly\footnote{https://github.com/CVI-SZU/Linly} and Firefly-LLaMA2-Chinese\footnote{https://github.com/yangjianxin1/Firefly-LLaMA2-Chinese} also conducted CPT practices on LLama-2 7B, 13B and 70B. Based upon the recent LlaMA-3~\cite{grattafiori2024llama3herdmodels}, Llama-3-Chinese conduct CPT on Llama-3 8B with LORA\footnote{https://github.com/ymcui/Chinese-LLaMA-Alpaca-3}, while Llama3-SynE~\cite{chen2024effectiveefficientcontinualpretraining} also perform CPT on 8B with synthetic data on scientific domains. 




For the second category of CPT, Domain-Adaptive Pre-training (DAPT)~\cite{gururangan2020dontstoppretrainingadapt} involves CPT on domain-specific corpora, and utilizes a corpus larger than task-specific datasets. EcomGPT-CT~\cite{ma2023ecomgptctcontinualpretrainingecommerce} performs CPT on Bloom on e-commerce semi-structured data. TRAIT~\cite{liang2024taskorientedindomaindata} adapt LLM to domains of advertisement and math by proposing a data-augmentation framework. Llama Pro-8.3B~\cite{wu2024llamaproprogressivellama} propose a relatively different methodology which post-trains Llama-2 7B with block expansion, for programming and mathematics domains.


In contrast, we enhance Llama-3 with Chinese corpus with not only the Chinese benchmarks but also some specialty domains improved. we have 1.7 trillion tokens of CPT corpus which is much larger than those of previous studies. We also conduct on Llama-70B while others might only study on 8B. 


\subsection{Scaling law of Pre-training and CPT}

After the forerunner efforts of scaling law on pre-training LLMs \cite{Kaplan2020OpenAIscalinglaw,Hoffmann2022DeepmindScalingLaw}, recently there have also been some attempts to discuss the scaling law of CPT. For example, D-CPT law \cite{que2024dcptlawdomainspecificcontinual} proposes the scaling law of loss with mixture ratio, dataset size, and model size, to study the loss contour on six domains. Their dataset sizes vary from 0.1 to 26B and model sizes vary from 0.5 to 4B. CMR \cite{gu2024cmrscalinglawpredicting} proposes a scaling law for CPT to two domains, finance and science QA, given a fixed training tokens budget. Training tokens are 220B and training model sizes vary from 460M to 3.1B. Furthermore, they define the term Critical Mixture Ratio which means the maximum data mixture ratio within its feasible region. In comparison, here we study the scaling law between LR and the mixture ratio and argue that the optimal mixture ratio should be determined from the optimal averaged benchmark performance, not only the pre-training loss. Furthermore, our experiments are conducted on the 8B model size, which to the best of our knowledge, is the largest among such types of studies.


\section{Methodology}
\label{sec:method}


Here we briefly introduce our implementation details, the baselines and the evaluation benchmarks.

\begin{table*}[h]
\caption{Details of Training Datasets.}
\label{tab:dataset_formal_names}
\centering
\setlength\tabcolsep{3pt}
\begin{tabular}{c|c|l|l}
\toprule
\multicolumn{1}{c|}{Stage} & \multicolumn{1}{c|}{Language} & \multicolumn{1}{c|}{Dataset} & \multicolumn{1}{c}{Domain} \\ 
\toprule
\multicolumn{1}{c|}{\multirow{5}[1]{*}{CPT}}    
& \multirow{1}[1]{*}{English} & Dolma & raw text \\
\cline{2-4}
& \multirow{1}[1]{*}{Chinese} & WanJuan, zhihu & raw text \\
\cline{2-4}
& \multirow{3}[1]{*}{multilingual} 
& AutoMathText, openwebmath, AMPS & math \\
&  & StackOverflow, starcoderdata & code \\
&  & proprietary & mixed \\
\cline{1-4}
\multicolumn{1}{c|}{\multirow{6}[2]{*}{SFT}}    
& \multirow{3}[1]{*}{English} & MuTual & multi-turn dialogue \\
&  & MathInstruct & mathematical QA \\
&  & Puffin & QA \\
\cline{2-4}
& \multirow{2}[1]{*}{Chinese} & crosswoz, Tiger-sft-zh & multi-turn dialogue \\
&  & stem-zh, COIG-CQIA & QA \\
\cline{2-4}
& \multirow{1}[1]{*}{multilingual} & alpaca & QA \\
\cline{1-4}
\multicolumn{1}{c|}{\multirow{3}[2]{*}{DPO}} 
& \multirow{1}[1]{*}{English} & PKU-SafeRLHF & human preference \\
\cline{2-4}
& \multirow{1}[1]{*}{Chinese} & CValues, huozi-rlhf & human preference \\
\cline{2-4}
& \multirow{1}[1]{*}{multilingual} & HC3 & human preference \\
\bottomrule
\end{tabular}

\end{table*}


\subsection{Implementation Details}


We employ the series of Llama-3~\cite{grattafiori2024llama3herdmodels} as our basis, which contains a limited fraction of Chinese tokens among their pre-training corpus. We conduct the pipeline of CPT, SFT and DPO starting from Llama-3 8B and 70B, respectively. We therefore name the resulting versions of LLM as CPT, C-SFT, and C-DPO, respectively. We also include results of Llama-3-Chinese~\cite{cui2023ChineseLLAMA} which also conducts CPT of Chinese corpus on Llama-3 8B. 



We run CPT in Megatron\footnote{https://github.com/NVIDIA/Megatron-LM} and finetuning in LlamaFactory~\cite{zheng2024llamafactory}. The 8B experiments are run by 32 A100 GPUs with 2 TP and 1 PP, while the 70B experiments are run by 128 GPUs with 8 TP and 2 PP. We use the AdamW optimizer with a sequence window length of 2048. Table~\ref{tab:dataset_formal_names} lists training datasets of three stages and Table~\ref{tab:congfiguration} lists some other configurations in our experiments.

\begin{table}[h!]
\caption{Experimental Configurations}
    \label{tab:congfiguration}
    \centering
\setlength\tabcolsep{10pt}
    \begin{tabular}{lccc}
        \toprule
        \multicolumn{1}{c}{Stage} & CPT & SFT & DPO \\ 
        \midrule
        micro batch size & 4 & 8 & 2 \\
        global batch size & 256 & 1024 & 128 \\
        LR scheduler & linear & cosine & cosine \\
        weight decay & 0.01 & 0.01 & 0 \\ 
        \bottomrule
    \end{tabular}
\end{table}


\subsection{Evaluation Benchmarks}

Here we briefly introduce the evaluation benchmarks we use in this work, categorized by their domains. We primarily use the Chinese benchmarks to evaluate the model capabilities in the newly adopted language, the English benchmark to ensure the robustness of general capability, and the reasoning, math, and coding benchmarks to indicate model improvement in domain-specific knowledge.

\paragraph{Disciplinary and Common Knowledge.} C-Eval~\cite{huang2023cevalmultilevelmultidisciplinechinese} is a set of Chinese multichoice questions (MCQ) covering different subjects and difficulty levels. MMLU~\cite{hendrycks2021measuringmassivemultitasklanguage} is an English MCQ benchmark in 57 subjects, including humanities, STEM, and social sciences. AGI-Eval~\cite{zhong2023agievalhumancentricbenchmarkevaluating} is a comprehensive English benchmark in various domains. TriviaQA~\cite{joshi2017triviaqalargescaledistantly} is a question-answering (QA) task consisting of trivia questions with corresponding answers.

\paragraph{Reasoning, Math and Coding.} BBH~\cite{srivastava2023imitationgamequantifyingextrapolating} is a set of tasks focusing on complex reasoning; GSM8K~\cite{cobbe2021trainingverifierssolvemath} is a set of 8.5K grade school math word problems. HumanEval~\cite{chen2021evaluatinglargelanguagemodels} evaluates code generation and program completion. 

\paragraph{Natural Language Understanding (NLU).} CMRC~\cite{Cui_2019} and DRCD~\cite{shao2019drcdchinesemachinereading} are reading \& understanding datasets in Chinese; CLUEWSC~\cite{xu2020cluechineselanguageunderstanding} is a Chinese version of the Winograd Schema Challenge; EPRSTMT~\cite{xu2021fewcluechinesefewshotlearning} is a Chinese sentiment analysis benchmark; LCSTS~\cite{hu-etal-2015-lcsts} is a Chinese short-text summarization benchmark; WMT22 zh$\rightarrow$en and en$\rightarrow$zh are the machine translation benchmarks between English and Chinese, originally published in 2022.

\paragraph{Comprehensive Alignment.} BelleEval~\cite{ji2023BelleExporeCases} is a comprehensive alignment benchmark with 1000 Chinese questions and corresponding answers.

\section{Study of Addition Language Mixture ratio}
\label{sec:almr}

Experiments of pre-training often last several weeks or months, therefore it is extremely difficult to find the theoretical optimal hyper-parameters. Similar cases are encountered for CPT studies although the training cost might be smaller. In this work, we focus on the case of CPT with extra language addition, which immediately calls for the key hyper-parameter, ALMR, a fraction of additional language corpus over the entire training corpus. In this work, we conduct different independent CPT experiments starting from Llama 8B, with different ALMR and LR combinations, to find out their optimal choices. Each training stops with nearly 100B tokens consumed when the training loss becomes stable.

\begin{figure*}[htbp]
    \centering
    \includegraphics[width=0.45\linewidth]{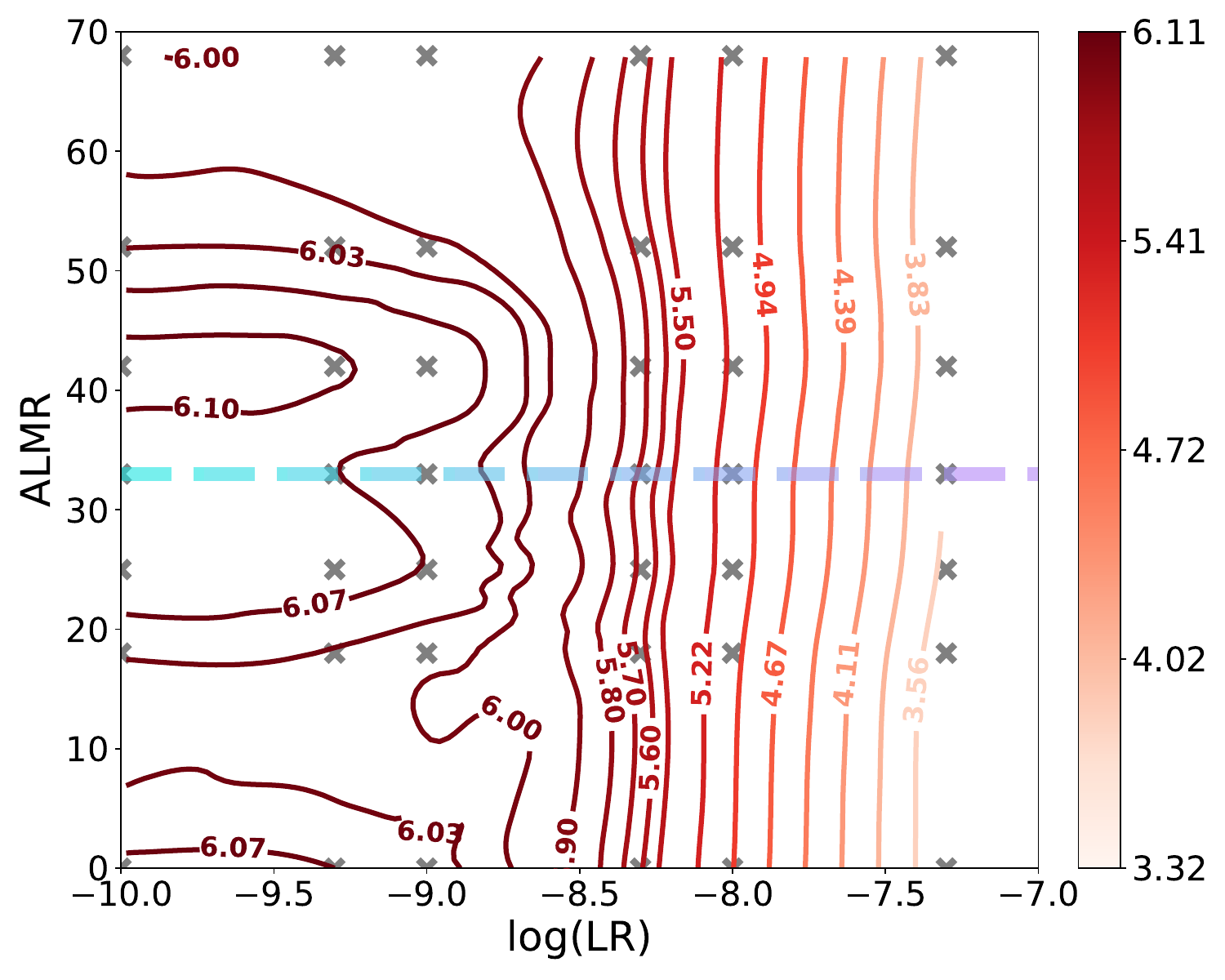}
    \hspace{0.1in}
    \includegraphics[width=0.45\linewidth]{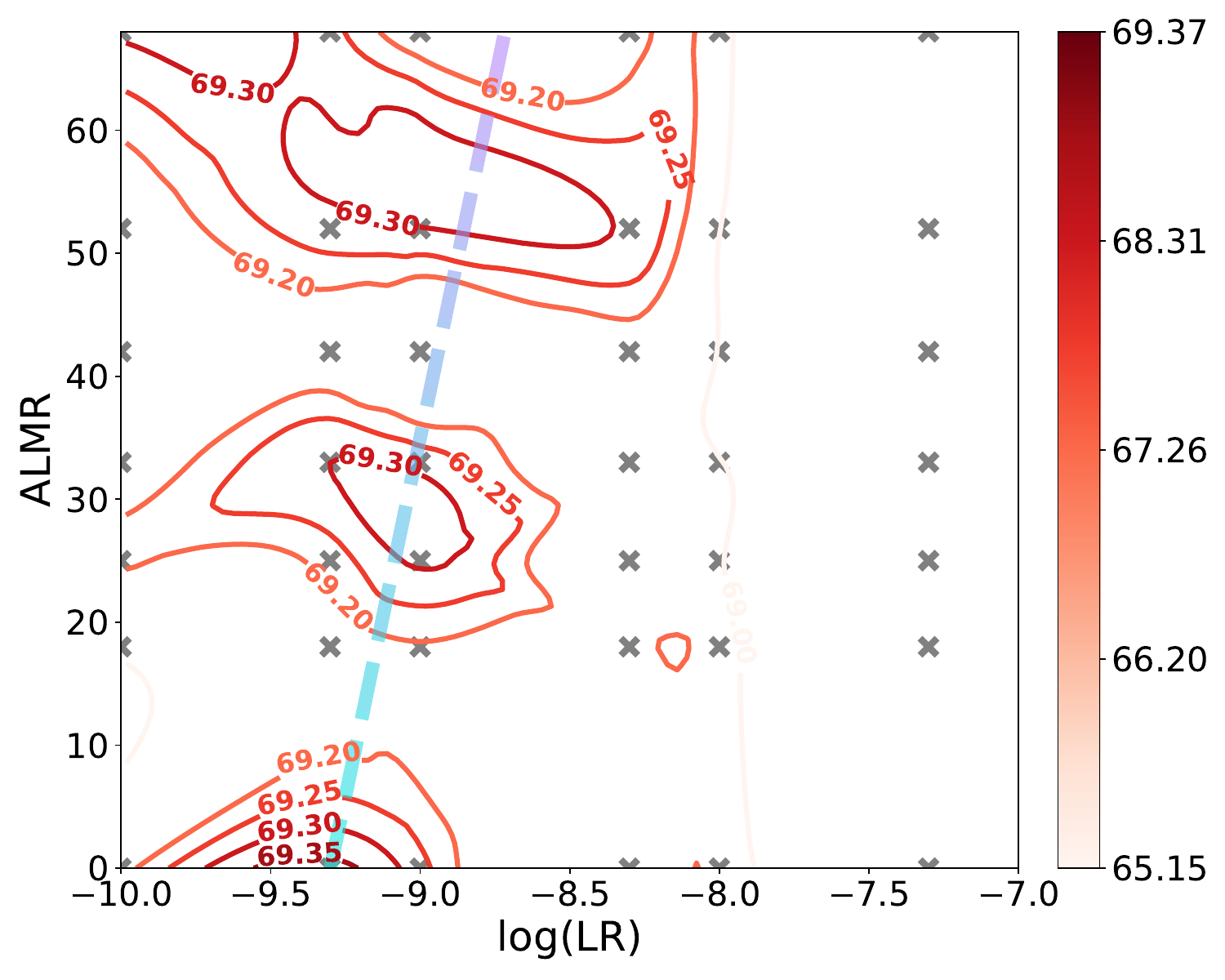}
    \caption{CPT Performance contours for different combinations of ALMR (in percentage) and LR on Llama-3 8B. The contour values correspond to validation loss (left) and averaged metrics (right). The cross points are experimental data points and the contours are extrapolated. The blue dash lines indicate the efficient frontiers between ALMR and LR found from the contours.}
    \label{fig:contour}
\end{figure*}

Fig.~\ref{fig:contour} depicts the contours of two important factors, validation loss and averaged metrics, as functions of ALMR and LR. The left part of Fig.~\ref{fig:contour} indicates that the larger LR is, the smaller the validation loss can reach. However, different from previous studies like D-CPT-law and CMR, here we argue that a small loss does not necessarily correspond to better performance (consider the case when there is no additional language or extra domain knowledge injected in CPT, then the loss would be small since LLM is already familiar with training tokens.). Therefore, we also present the right part of Fig.~\ref{fig:contour}, in which we average the benchmarks listed in Table~\ref{tab:pretrain_benchmarks} and plot the contour. Apparently, a larger mean metric means better performance, which corresponds to several peaks on the contours. We calculate their performance peaks and regress them to obtain the efficient frontier, which is the dashed line in the right sub-figure of Figure \ref{fig:contour}
\begin{equation}
\label{eq:metric_lr_almr}
\text{ALMR} = 116.67 \log (\text{LR}) + 1085.00
\end{equation}

To determine the final choice of our ALMR and LR in the final experiment, we also draw the efficient frontier Figure \ref{fig:contour} (left), by calculating the gradient of LR and ALMR along which the loss decreases fastest
\begin{equation}
\label{eq:loss_lr_almr}
\text{ALMR} = -0.33 \log (\text{LR}) + 29.67
\end{equation}

Equation \ref{eq:metric_lr_almr} and Equation \ref{eq:loss_lr_almr} intersect at ALMR = $33\%$ and LR = $1.0e-9$, which are the formal choices of our ALMR and LR, in the 8B experiment. For the 70B experiment, we decrease LR to 1.0e-10 and adjust AMLR accordingly.



\section{Experimental Results}
\label{sec:experiment}

In this section, we exhibit the results of CPT, SFT and DPO studies. We name the latter two \textbf{C-SFT} and \textbf{C-DPO}, respectively, as they are further finetuned versions of \textbf{CPT}. We also conduct ablation studies and further alignment to enhance emotional intelligence. Experimental results indicate that CPT results in substantial improvement of downstream task performances.

\begin{figure*}[htbp]
    \centering
    \includegraphics[width=0.48\linewidth]{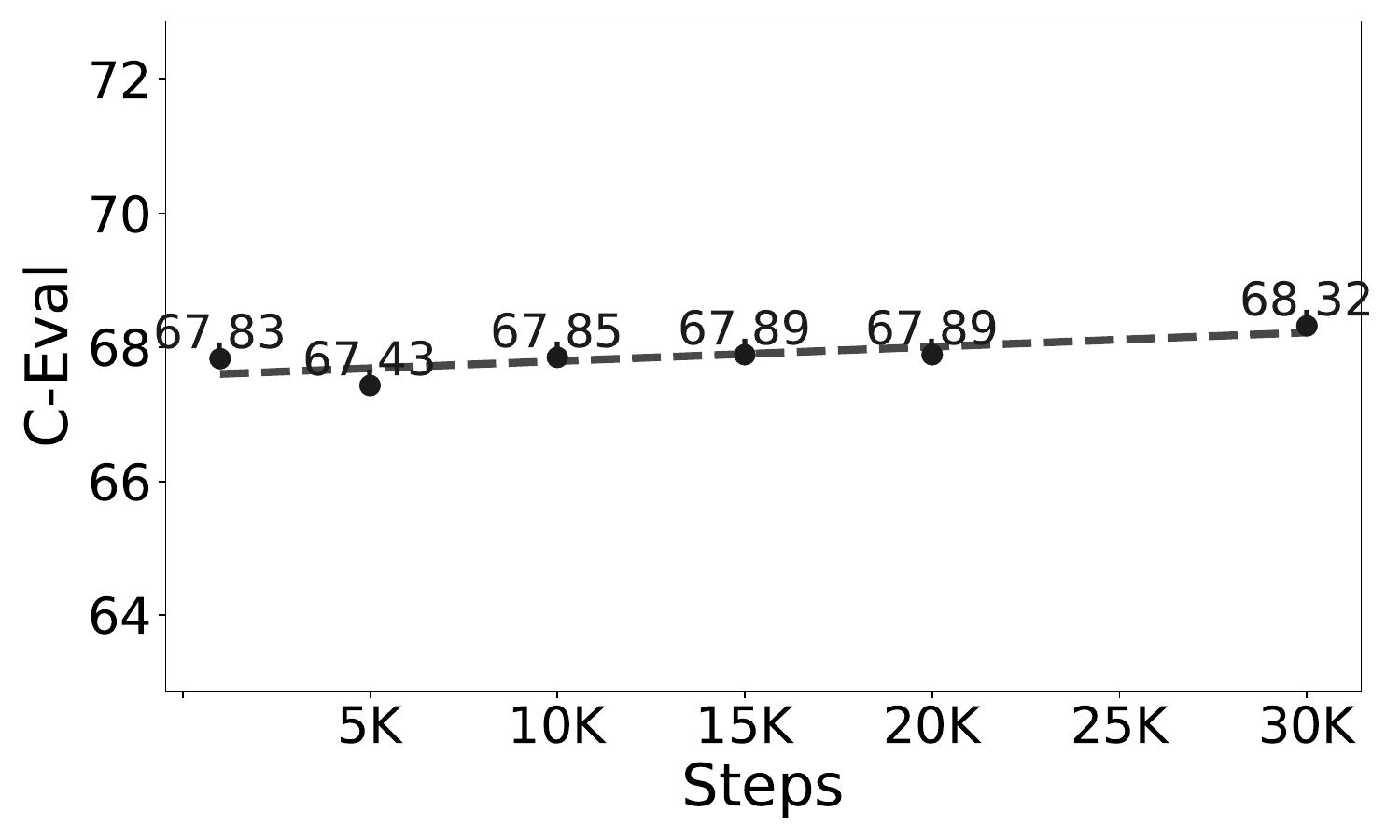}
    \hspace{0.1in}
    \includegraphics[width=0.48\linewidth]{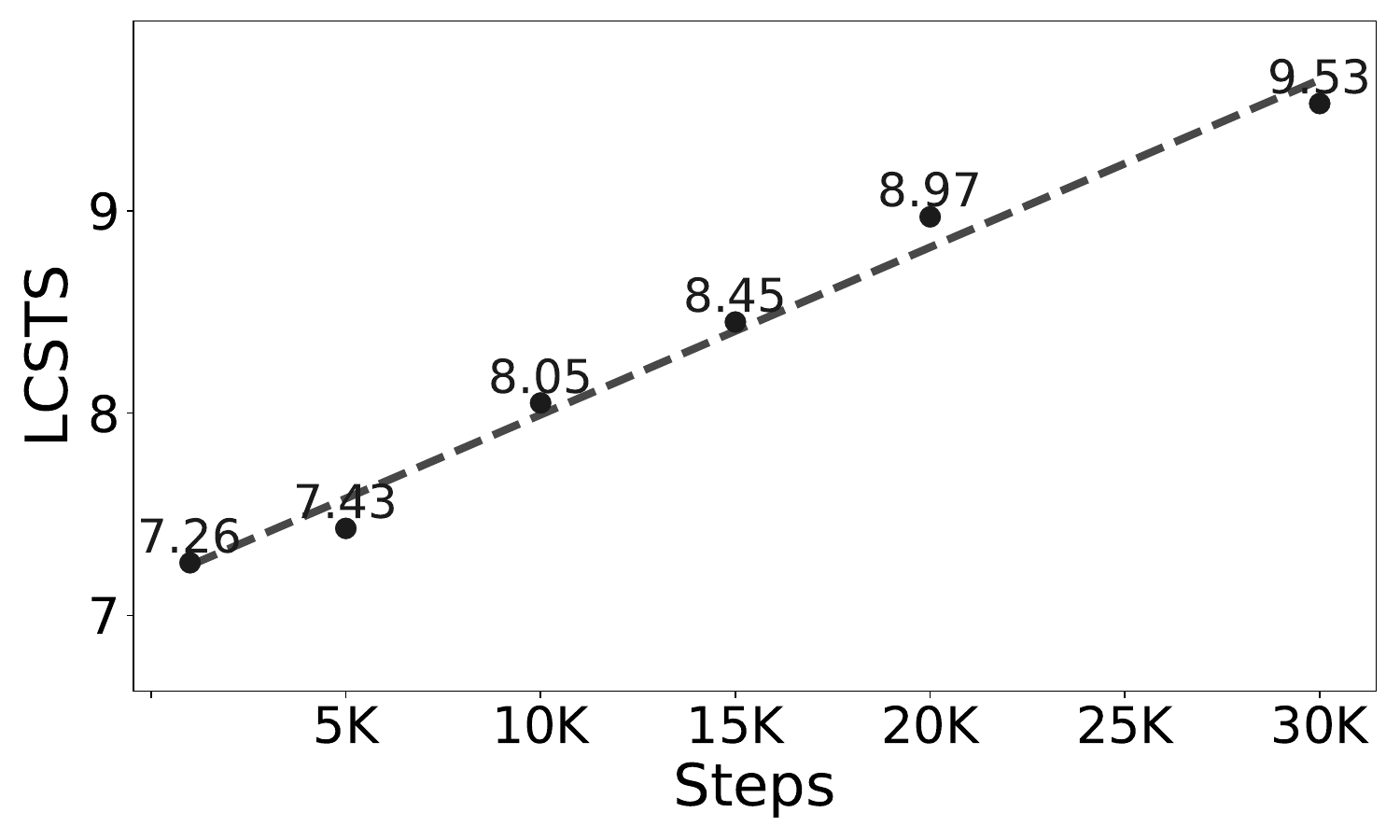}

    \includegraphics[width=0.48\linewidth]{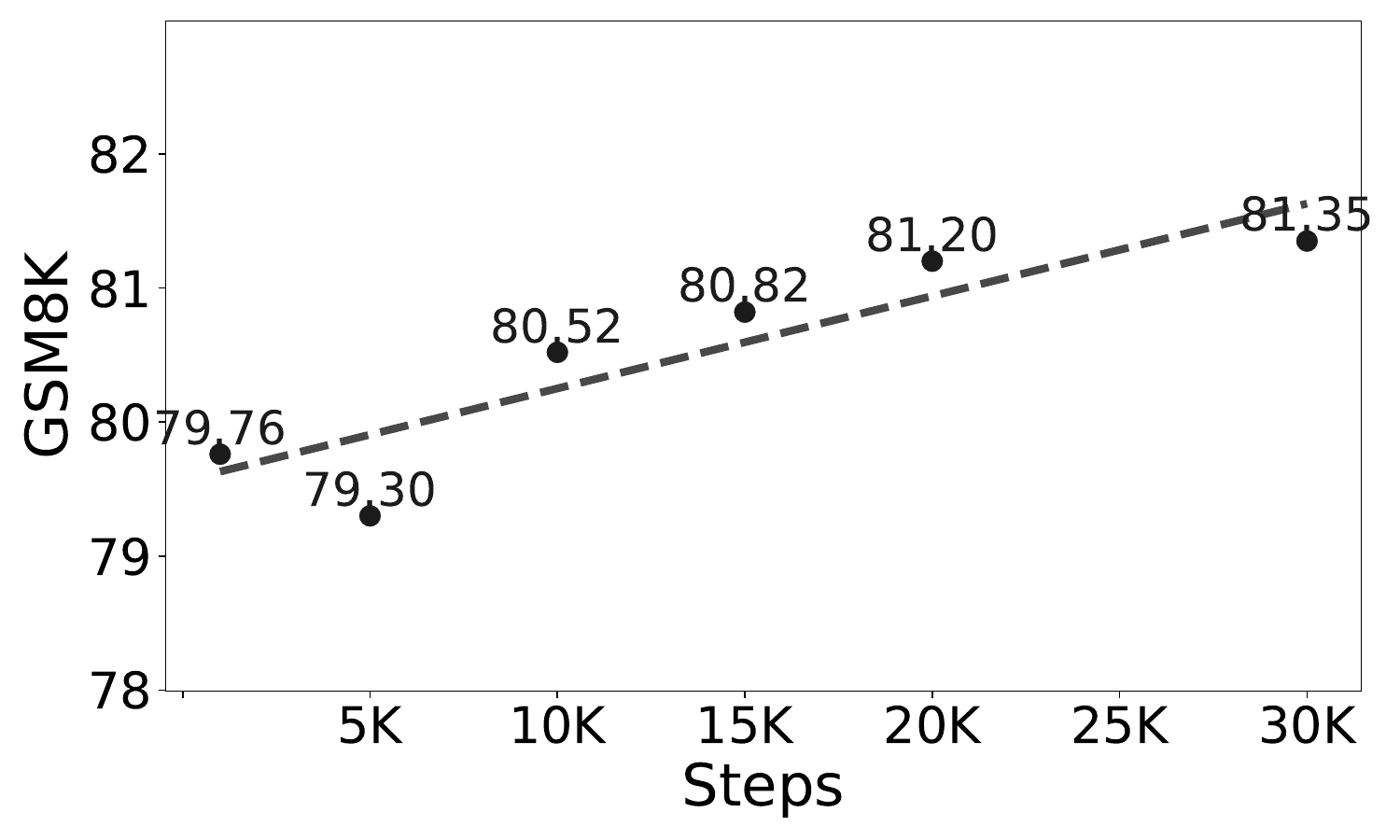}
    \hspace{0.1in}
    \includegraphics[width=0.48\linewidth]{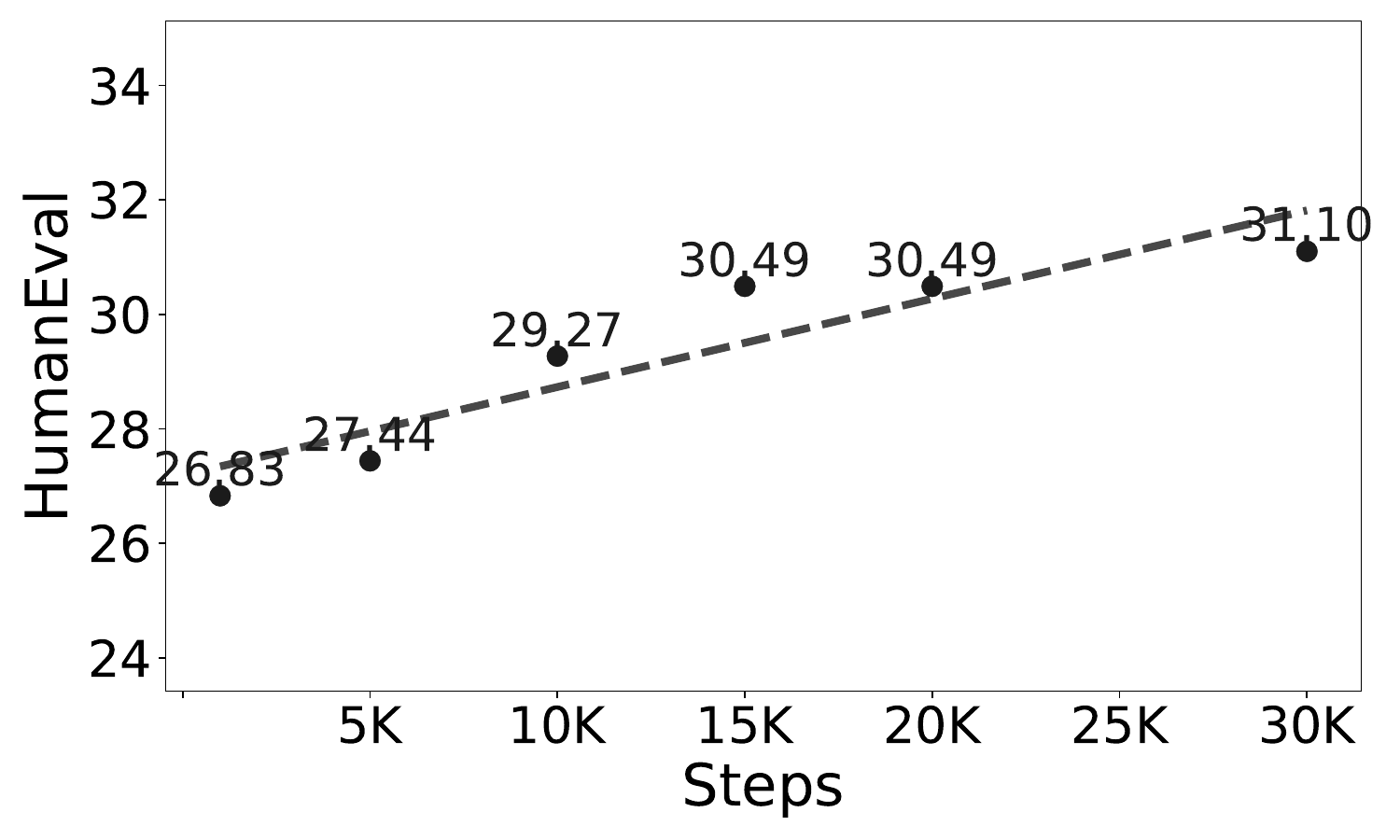}
    
    \caption{Typical metric plots of CPT experiment on Llama-3 70B. Metrics include C-Eval, LCSTS, GSM8K and HumanEval.}
    \label{fig:metric_curves}
\end{figure*}

\subsection{Results of Continual Pre-Training} 

We mainly employ the benchmarks of disciplinary, common, and specific domain knowledge, to evaluate the capability of pre-trained models. Table~\ref{tab:pretrain_benchmarks} lists the results of the aforementioned benchmarks, in which we compare our CPT model with the original Llama-3, as well as Llama-3-Chinese 8Be~\cite{cui2023ChineseLLAMA}. For 8B comparison, we achieved better results in all Chinese, Reasoning, Math and Coding benchmarks, while the results of English benchmarks remained almost the same. It is also worthwhile to notice that Llama-3-Chinese fails to maintain the general and domain skills, with MMLU, GSM8K and HumanEval all apparently decreased from Base. Furthermore, our 70B CPT model outperforms Llama-3 in almost all the benchmarks, with the only exception of BBH.

\begin{table*}[htbp]
\caption{Evaluated Benchmarks of Pre-training Models. C-Eval and MMLU are 5-shot; BBH is 3-shot; GSM8K is 4-shot; AGI-Eval and HumanEval are zero-shot. Results of Llama-3-Chinese are originally reported by the authors (\url{https://github.com/ymcui/Chinese-LLaMA-Alpaca-3}).}
\label{tab:pretrain_benchmarks}
\begin{center}
\scriptsize
\setlength\tabcolsep{3pt}
\begin{tabular}{ccccccccc}
\toprule
\multirow{2}{*}{Size}  & \multirow{2}{*}{Model} 
& \multicolumn{2}{c}{Chinese} & \multicolumn{2}{c}{English} & \multicolumn{3}{c}{Reasoning, Math and Coding} \\
\cmidrule(lr){3-4} \cmidrule(lr){5-6} \cmidrule(lr){7-9}
& & C-Eval & LCSTS & MMLU & AGI-Eval & BBH & GSM8K & HumanEval \\
\toprule
\multirow{ 3}{*}{8B}    
& Llama-3   & 49.38 & 9.64 & \bf 66.88  & \bf 37.55 & 58.50 & 54.44 & 26.22 \\
& Llama-3-Chinese & 50.50 & - & 61.10  & - & - & 37.98 & 9.76 \\
& \bf CPT (ours)   & \bf 51.12 & \bf 9.76 & 66.87  & 37.44 & \bf 58.87 & \bf 54.89 & \bf 26.83 \\
\midrule
\multirow{ 2}{*}{70B}    
& Llama-3   & 67.75 & 7.19 & 79.47  & 50.83 & \bf 77.62 & 79.38 & 28.05 \\
& \bf CPT (ours)    & \bf 68.36 & \bf 8.97 & \bf 79.53  & \bf 51.05 & 77.56 & \bf 81.40 & \bf 31.23 \\
\bottomrule
\end{tabular}
\end{center}

\end{table*}


We also visualize several typical metric curves, to show the tendency when the training dataset can be even expanded. Fig.~\ref{fig:metric_curves} shows the time trajectories of C-Eval, LCSTS, GSM8K and HumanEval. From the figure, we speculate that disciplinary knowledge can be relatively quickly obtained but then improved slower by CPT, while NLU and domain knowledge are consistently improved along the entire CPT stage, and can be further improved given more high-quality data. We note that this conclusion is similar to some previous research~\cite{ji2023BelleExporeCases}.


\begin{table*}[h!]
\caption{Evaluated Benchmarks of Fine-tuned Models. WMT22 zh$\rightarrow$en and WMT22 en$\rightarrow$zh are 5-shot; all other benchmarks are zero-shot. We use `Instruct' to abbreviate Llama-3 Instruct of 8B or 70B; C-SFT and C-DPO mean SFT and DPO models trained from CPT checkpoints.}
\label{tab:finetune_benchmarks}

\begin{center}
\scriptsize
\begin{tabular}{ccccccccccc}
\toprule
\multirow{2}{*}{\bf Size}  & \multirow{2}{*}{Model} 
& \multicolumn{6}{c}{Chinese} & \multicolumn{1}{c}{English} & \multicolumn{2}{c}{WMT22} \\ 
\cmidrule(lr){3-8} \cmidrule(lr){9-9} \cmidrule(lr){10-11} 
& & LCSTS & CLUEWSC & CMRC & DRCD & EPRSTMT & BelleEval & TriviaQA &  zh$\rightarrow$en & en$\rightarrow$zh \\
\toprule
\multirow{ 3}{*}{8B}    
& Instruct   & 6.04 & 80.12 & 87.69  & 90.83 & 83.11 & 0.74 & 65.29 & 21.83 & 22.07 \\
& \bf C-SFT   & \bf 14.01 & 80.64 & 92.26  & 90.32 & \bf 88.52 & 0.79 & 65.02 & \bf 26.49 & \bf 24.95 \\
& \bf C-DPO    & 11.19 & \bf 85.04 & \bf 93.48  & \bf 92.71 & 87.70 & \bf 0.81 & \bf 66.49 & 25.74 & 24.89 \\
\midrule
\multirow{ 3}{*}{70B}    
& Instruct   & 7.12 & 86.37 & 91.21  & 95.09 & 89.02 & 0.78 & 79.88 & 28.03 & 24.05 \\
& \bf C-SFT    & \bf 15.00 & 90.88 & 96.52  & 96.51 & \bf 91.15 & 0.81 & \bf 81.51 & 32.41 & 29.17 \\
& \bf C-DPO    & 12.40 & \bf 90.98 & \bf 96.71  & \bf 96.88 & 90.82 & \bf 0.87 & 81.03 & \bf 33.01 & \bf 29.49 \\
\bottomrule
\end{tabular}
\end{center}

\end{table*}

\subsection{Results of Alignment}

Table~\ref{tab:finetune_benchmarks} lists the results of C-SFT and C-DPO, with a comparison with the official Llama-3 instruct models. Again, our models outperform the baselines in all benchmarks, where C-SFT performs better in some benchmarks while C-DPO performs better in others. We argue that the preference data used by C-DPO would steer LLM to its interested domain, in the cost of degradation of some other knowledge.


Note that the LCSTS results of Llama-3-instruct in Table~\ref{tab:finetune_benchmarks} have apparently been improved compared to those in Table~\ref{tab:degradation}. The reason is that here we append a prompt like `Please summarize in Chinese.' to the end of LCSTS original prompts, to encourage Llama-3-instruct to respond in Chinese therefore the performance comparison is more meaningful. 




\begin{table}[ht]
    \caption{Ablation study of on our C-DPO 8B. `finetune-w-cn' means if including the Chinese samples during SFT and DPO stages.}
    \label{tab:ablation}
    \centering
    \setlength\tabcolsep{4pt}
    \begin{tabular}{ccccc}
        \toprule
        \multirow{1}{*}{CPT} & \multirow{1}{*}{finetune-w-cn} & 
         LCSTS & CLUEWSC & EPRSTMT \\
        \midrule
        $\times$ & \checkmark & 10.59 & 76.33 & 85.08 \\
        \checkmark & $\times$ & 4.89 & 56.25 & 83.11 \\
        \checkmark & \checkmark & \bf 11.19 & \bf 85.04 & \bf 87.70  \\
        \bottomrule
    \end{tabular}

\end{table}

\subsection{Ablation Study}

To further validate the effectiveness of our methodology, we conduct the ablation study on C-DPO 8B, to compare the performances with or without the CPT stage, and with or without the Chinese corpus during fine-tuning stages. We list the results of LCSTS, CLUEWSC and EPRSTMT  in Table~\ref{tab:ablation}. It is validated that C-DPO still performs the best, and both CPT and fine-tuning with Chinese samples are necessary, contributing to the final overall performance.

\begin{table*}[h!]
\caption{Emotional Intelligence Performances of Finetuned Models. `sent' is the abbreviation of sentiment, and smaller sent values indicate better results. EQ represents the emotional quotient. `cn' and `en' represent Chinese and English, respectively.}
\label{tab:psychology_results}
\begin{center}
\scriptsize
\setlength\tabcolsep{4pt}
\begin{tabular}{ccccccc}
\toprule
\multirow{2}{*}{Model} & \multirow{2}{*}{TAPTAP $\uparrow$}  & \multirow{2}{*}{Turing Test $\uparrow$} 
& \multicolumn{4}{c}{SECEU} \\
\cmidrule(lr){4-7}
& & & sent-cn $\downarrow$ & sent-en $\downarrow$ & EQ-cn $\uparrow$ & EQ-en $\uparrow$ \\
\midrule
Qwen2-72B-Instruct    & 72.80 & 5.11 & 2.21  & 2.37 & 110.60 & 107.70 \\
Llama3-70B-Instruct   & 45.80 & 13.60 & 2.36  & 2.64 & 107.80 & 102.76 \\
C-SFT (ours)   & 76.80 & 9.23 & 1.94  & 2.43 & 115.49 & 106.57 \\
C-SFT-Empathy (ours) & \bf 78.20 & \bf 14.16 & \bf 1.83  & \bf 2.27 & \bf 117.59 & \bf 109.50 \\
\bottomrule
\end{tabular}
\end{center}

\end{table*}

\subsection{Deployment on an Empathetic Chatting System}


The ultimate purpose of this project is to deploy the 70B Chinese-enhanced LLM on a non-commercial chatbot application. To build an empathetic companion, we further finetune C-SFT 70B with emotional supporting conversation (ESC) samples, in both English and Chinese. We call this empathy-strengthened version by \textbf{C-SFT-Empathy}.

To evaluate its emotional intelligence, we examine the emotion classification benchmark TAPTAP~\cite{zhang2023generativetablepretrainingempowers} and the emotional intelligence benchmark SECEU~\cite{SECEU}, as well as conduct the famous Turing Test\footnote{We follow the evaluation approach proposed by BotChat~\cite{duan2023botchatevaluatingllmscapabilities} and use 500 multi-turn proprietary dialogues as the test test.}. We list the results of the above benchmarks in Table~\ref{tab:psychology_results}, with the baselines of Llama3-70B-instruct and also Qwen2-72B-Instruct~\cite{qwen2techreport2023}. Results show that C-SFT-Empathy surpasses baselines in all benchmarks and C-SFT performs the second.




Based on results in Table~\ref{tab:psychology_results}, we choose C-SFT-Empathy as the final deploying model. We collect online conversation data for one week, and compare it with the responses of Llama3-70B-Instruct and GPT4 upon the same query. To examine the response qualities, pairwise evaluations are conducted by human annotators among three models, with the win rates shown in Table~\ref{tab:chat_result}. Win rate results indicate that C-SFT-Empathy has better performance than Llama3-70B-Instruct and GPT4, for Chinese conversation scenarios. We also calculate the fraction of Chinese tokens among model responses, as shown by \%Chinese in Table~\ref{tab:chat_result}. Again, Llama3-70B-Instruct fails to align with the Chinese query and often responds in other languages, with only a 9\% percentage of Chinese tokens in responses.

\begin{table}[h!]
\caption{Pairwise evaluated win rates between ours C-SFT-Empathy 70B, Llama3-70B-Instruct and GPT4, as well as percentages of Chinese tokens over online responses.} 
    \label{tab:chat_result}
    \centering
    \small
    \setlength\tabcolsep{5pt}
    \begin{tabular}{ccc|c}
    \toprule
    Model & vs Llama3-70B-Instruct & vs GPT4 & \%Chinese \\ 
    \midrule
    Llama3-70B-Instruct & N/A & 10.2 & 9.0 \\
    C-SFT-Empathy (ours) & \bf 97.5 & \bf 79.4 & \bf 77.3 \\ 
    \bottomrule
    \end{tabular}
\end{table}



\section{Conclusion}
\label{sec:conclusion}

In this paper, we conduct post-traing including CPT, SFT, and DPO on Llama-3 8B and 70B, on the purpose of strengthening Llama on a specific additional language. Surprisingly, we find that the CPT of extra language also boosts LLM's capability to understand not only additional language but also some other specific domains. We also successfully finetuned it with more dialogue-like samples and obtained state-of-the-art performance on emotional intelligence benchmarks. This version of the model has been deployed on an industrial-scale application and provides real-life emotional chatting support. We also conduct substantial experiments on the optimal correlation between the learning rate and additional language mixture ratio, which sheds some lights on future CPT researches.

\bibliographystyle{splncs04}
\bibliography{paper-472}
%

\end{document}